\documentclass[letterpaper, 10 pt, conference]{ieeeconf}  

\IEEEoverridecommandlockouts                              

\usepackage{graphics} 
\usepackage{mathptmx} 
\usepackage{amsmath} 
\usepackage{amssymb}  
\usepackage{pdfpages}
\usepackage[utf8]{inputenc} 
\usepackage{caption}
\usepackage{subcaption}
\usepackage{algpseudocode}
\usepackage{algorithm}
\usepackage{booktabs}
\usepackage{multirow}
\usepackage{todonotes}
\usepackage{mathrsfs}
\usepackage{tcolorbox}
\usepackage{fancyhdr} 
\usepackage{background}

\backgroundsetup{
  scale=1,
  color=black,
  opacity=1,
  angle=0,
  position=current page.north,
  vshift=-0.5cm,
  hshift=0cm,
  contents={%
    \begin{tcolorbox}[colframe=white, colback=white, boxrule=0mm, sharp corners, width=\textwidth, enlarge left by=0cm, enlarge right by=0cm]
    \centering
    \textbf{This paper has been accepted by IEEE ITSC 2024.}
    \end{tcolorbox}
  }
}

\title{\LARGE \bf
Safety Metric Aware Trajectory Repairing for Automated Driving
}

\author{Kailin Tong$^{1}$, Berin Dikic$^{1}$, Wenbo Xiao$^{2}$, Martin Steinberger$^{3}$, Martin Horn$^{3}$ and Selim Solmaz$^{1}$
	\thanks{* The work was supported by the project ARCHIMEDES, which is Co-funded by the European Union and by the Chips Joint Undertaking and its members including top-up funding of the Austrian Federal Ministry for Climate Action (BMK) under Grant Agreement No 101112295. Views and opinions expressed are however those of the author(s) only and do not necessarily reflect those of the European Union Key Digital Technologies Joint Undertaking. Neither the European Union nor the granting authority can be held responsible for them.}
	\thanks{$^{1}$ K. Tong, B. Dikic and S. Solmaz are with Virtual Vehicle Research GmbH, Inffeldgasse 21a, 8010 Graz, Austria. {\tt\small \{kailin.tong, berin.dikic, selim.solmaz\}@v2c2.at}
	} 
    \thanks{$^{2}$ W. Xiao is with the Institute of Automotive Engineering at Graz University of Technology, Inffeldgasse 11, 8010 Graz, Austria. {\tt\small wenbo.xiao@student.tugraz.at}
    } 
	\thanks{$^{3}$ M. Steinberger and M. Horn are with the Institute of Automation and Control at Graz University of Technology, Inffeldgasse 21b, 8010 Graz, Austria.
	{\tt\small  \{martin.steinberger, martin.horn\}@tugraz.at}
	}
}
\begin{document}
\maketitle
\thispagestyle{empty}
\pagestyle{empty}

\begin{abstract}
Recent analyses highlight challenges in autonomous vehicle technologies, particularly failures in decision-making under dynamic or emergency conditions. Traditional automated driving systems recalculate the entire trajectory in a changing environment. Instead, a novel approach retains valid trajectory segments, minimizing the need for complete replanning and reducing changes to the original plan. This work introduces a trajectory repairing framework that calculates a feasible evasive trajectory while computing the Feasible Time-to-React (F-TTR), balancing the maintenance of the original plan with safety assurance. The framework employs a binary search algorithm to iteratively create repaired trajectories, guaranteeing both the safety and feasibility of the trajectory repairing result. In contrast to earlier approaches that separated the calculation of safety metrics from trajectory repairing, which resulted in unsuccessful plans for evasive maneuvers, our work has the anytime capability to provide both a Feasible Time-to-React and an evasive trajectory for further execution.
\end{abstract}

\section{Introduction}
Recent analysis of accident data \cite{mccarthy2022autonomous} underscores a persistent challenge in the development of autonomous vehicles (AVs): occasional failure to make optimal decisions, resulting in potential property damage or injuries, especially in emergency scenarios. Despite rigorous testing, the dynamic nature of traffic can abruptly alter the behavior of other vehicles, creating hazardous conditions. Automated driving systems typically respond by recalculating trajectories from the current state to the desired destination. However, this continual search for alternative trajectories is resource-intensive. A more efficient strategy, proposed in recent research \cite{tong2023robust}, involves identifying and retaining segments of a trajectory that remain valid, thus avoiding the need for wholesale re-planning. By selectively repairing only the affected portions, this approach minimizes computational overhead and enhances resilience to minor disturbances. 

Based on our previous knowledge of the use of quadratic programming for trajectory repairing \cite{tong2023robust} and the real-life demonstration of pedestrian collision avoidance systems \cite{Schratter.2019}, we have observed a distinct trade-off between replanning and repairing at a critical point. Making early changes to the original plan can result in a smoother reaction, but it might also significantly alter the intended trajectory, which is not always necessary in a dynamic environment. Performing crucial adjustments prevents possible accidents at the last moment and enables adherence to the original plan if possible. However, this approach may include excessively aggressive and evasive maneuvers. The balance between re-planning and critical repairing creates an optimization challenge that has been largely overlooked.

\begin{figure}[t] \label{fig:concept}
     \centering
     \begin{subfigure}[b]{0.9\columnwidth}
         \centering
          \includegraphics[width=1.0\columnwidth]{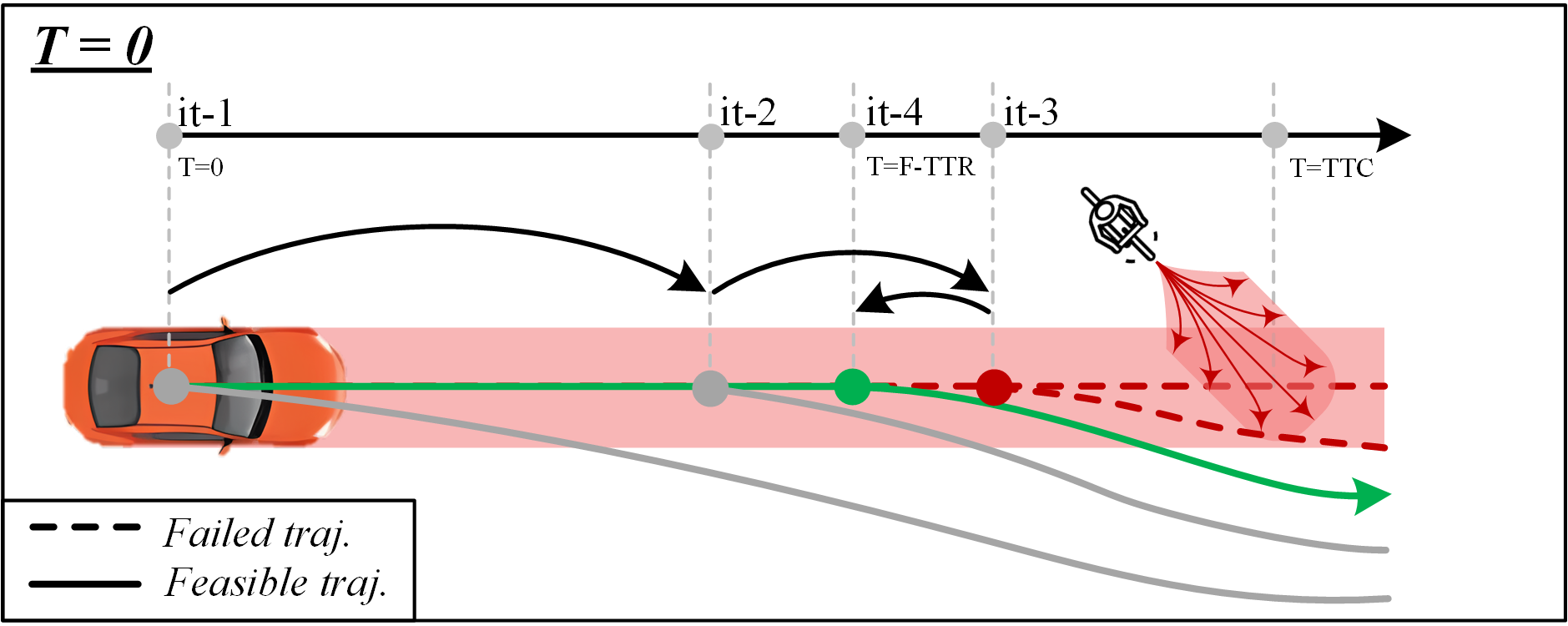}
         \caption{Trajectory repairing is performed when a conflict occurs between the cyclist and the ego vehicle. The result of iteration 4 is chosen as the repaired trajectory by binary search.}
         \label{fig:repair_t0}
     \end{subfigure}
     \hfill
     \begin{subfigure}[b]{0.9\columnwidth}
         \centering
          \includegraphics[width=1.0\columnwidth]{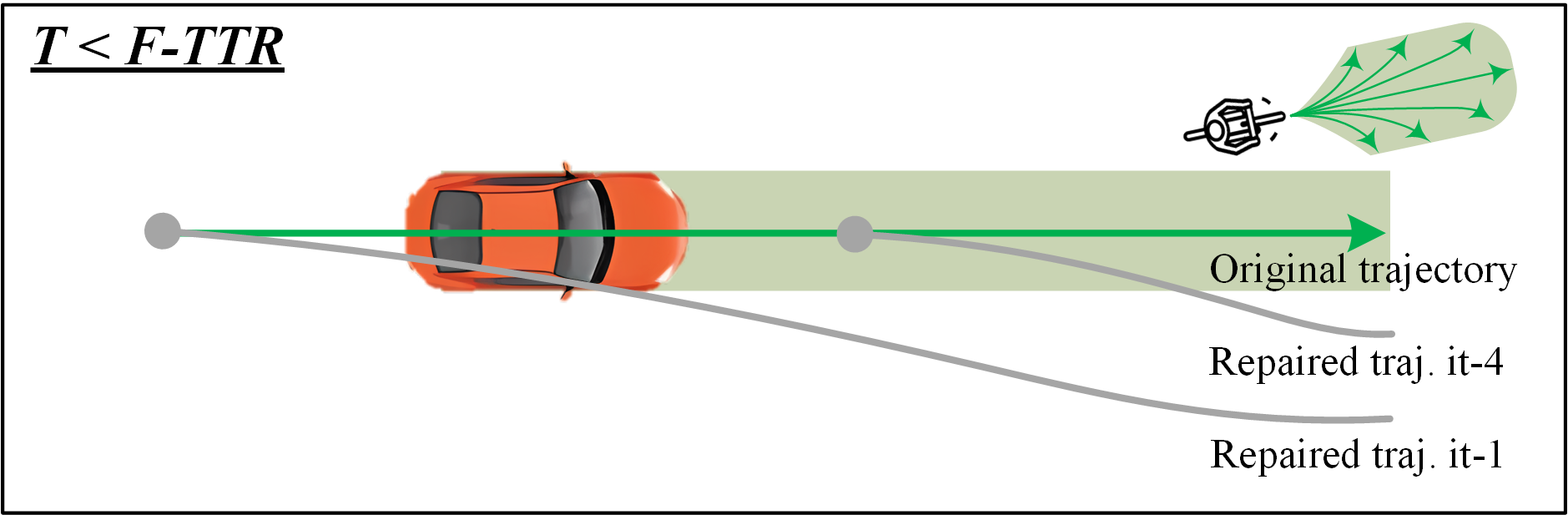}
         \caption{The ego vehicle continues on the original trajectory because the conflict disappears before the start of the repaired trajectory from the Feasible Time-to-React (F-TTR).}
         \label{fig:repair_trep}
     \end{subfigure}
     \hfill
    \caption{Trajectory repairing while computing safety metrics. 
}
\end{figure} 

A motivating example of our work is collision avoidance with a vulnerable road user (VRU), as illustrated in Fig. \ref{fig:repair_t0}, where a cyclist approaches the road curb and appears poised to cross. Typically, AVs would execute an immediate evasive maneuver, following a re-planned trajectory from $T=0$. However, our more robust approach involves identifying the Feasible Time-To-React (F-TTR), which not only ensures safety but also allows the system time to verify the VRU's future motion. As shown in Fig. \ref{fig:repair_trep}, this method provides the AV the opportunity to continue on its original trajectory if the cyclist alters the intended path.

This research introduces a trajectory repairing framework that is attuned to safety metrics, aimed at addressing the identified challenges. We strive to pinpoint the most critical but still executable intervention point for the trajectory planner (the green point in Fig. \ref{fig:repair_t0}) and generate a feasible trajectory from it. The binary search method used in our iterative approach is visualized in Fig. \ref{fig:repair_t0}, spanning iterations one through four.

The primary contributions of this research are summarized as follows:
\begin{itemize}
    \item We developed a trajectory repairing framework that provides a feasible evasive trajectory while incorporating safety metrics. Unlike previous studies that used simplified planners to estimate the Time-to-React (TTR), our framework utilizes the same planner both for approximating safety metrics and for actual maneuver, ensuring that the trajectories are feasible, collision-free and can be executed by the low-level controller.
    \item We introduced the concept of the Feasible Time-to-React (F-TTR), which serves as a practical approximation of the Time-to-React (TTR) when a feasible and collision-free trajectory is available for execution. This approach effectively bridges the gap between the theoretically calculated TTR values and the actual implementation of evasive maneuvers based on those timings. F-TTR addresses the limitations of traditional trajectory repairing planners that struggle to generate feasible maneuvers when initialized with TTR values derived from simplistic models. This advancement ensures more reliable and practical responses in dynamic driving scenarios.
\end{itemize}

The remainder of this paper is structured as follows: In Section \ref{sec:SotA}, we explore related work, examining safety metrics and trajectory repairing for autonomous systems. Section \ref{sec:Prelim} covers the preliminaries, discussing the vehicle model and configuration space. Following this, Section \ref{sec:Method} presents the solution method for trajectory repairing based on specific safety metrics. Subsequently, Section \ref{sec:eval} provides a detailed evaluation of simulation results, leading to conclusions and a discussion of future work in Section \ref{sec:conclusions}.

\section{Related Work}\label{sec:SotA}
\subsection{Computing Safety Metrics for Automated Driving}

One widely adopted safety assessment method in automated driving systems (AD) is Time-To-X (TTX) metrics, where X denotes various critical collision reactions. For instance, Time-to-Collision (TTC) evaluates the duration until a potential collision and triggers alerts or interventions \cite{Guo.2020}. Additional metrics like Time-To-Brake (TTB), Time-To-Kickdown (TTK), and Time-To-Steer (TTS) provide insights into braking, acceleration, and steering dynamics. Moreover, the Time-To-React (TTR) integrates these metrics for comprehensive worst-case scenario assessment \cite{Hillenbrand}. 

TTX calculation can be conducted online through empirical estimation or forward simulation methods. Schratter et al. utilize an empirical formula to estimate TTB and TTS, evaluating the collision risk for emergency maneuver decision-making \cite{Schratter.2019}. However, extending this approach to diverse scenarios poses challenges.

In forward simulation, reachable set analysis and modified binary search algorithms are employed for TTR computation \cite{sontges2018worst, lin2021sampling, tong2023robust}. This approach allows for the accurate computation of TTX values through simulation. However, incorporating longitudinal and lateral emergency maneuvers may not align with typical driver behavior and could be disfavored. Moreover, the disparity between the trajectory used for safety metric estimation and that for executing an evasive maneuver may lead to infeasible or sub-optimal results.

Addressing the need for comprehensive tools in the assessment of autonomous vehicle safety, Lin and Althoff \cite{lin2023crime} introduced CommonRoad-CriMe, an open-source toolbox for assessing autonomous vehicle safety. This tool offers a comprehensive coverage of criticality measures, facilitating their utilization and evaluation across various traffic scenarios. With visualized information for debugging and presentation, along with support for numerical experiments, CommonRoad-CriMe enables an efficient comparison of criticality measures and analysis of traffic conflicts.

\subsection{Trajectory Repairing}

Unlike re-planning, repairing means that only the necessary part of a reference trajectory is changed due to environmental disturbances. This concept has been widely used in the robotics domain as well as for Unmanned Aerial Vehicles (UAVs), and Unmanned Ground Vehicles (UGVs), in the form of local planning \cite{Zhou.2019, EGO2021Zhou}. 
However, they are not specifically designed for AVs and may not necessarily provide a safety metric. The initialization of ``repairing'' is contingent upon the specific configuration of optimization. Lin et al. \cite{lin2021sampling} introduced a trajectory repairing approach that utilizes closed-loop rapidly-exploring random trees (CL-RRT). They also devised a safety assurance mechanism for the generated evasive maneuver. Tong et al. proposed a path-speed decoupled trajectory repairing framework \cite{tong2023robust}. Additionally, a robustness measure ($\alpha$) are introduced in their work for fine-tuning the driving behavior. 
Furthermore, the improvement of trajectory repairing is achieved by the exploration of the satisfiability modulo theories paradigm \cite{Lin2022Rule}.

In the current state-of-the-art, safety metric computation and trajectory repairing are separate processes. Typically, a simplistic planner (e.g., evasive steering to the left) estimates the safety metric, and then a more sophisticated planner creates a trajectory for the low-level controller to follow. Our work bridges this gap by integrating trajectory repairing into the safety metric (F-TTR in our work) computation via binary searching. Thus, the trajectory planned during the safety metric assessment is the same one the low-level controller will track, avoiding the inefficiencies and even infeasible results of using two separate planners in prior approaches.

\section{Preliminaries}\label{sec:Prelim}
\subsection{Vehicle Model and Configuration Space} \label{sec:vehicle model}

In this study, we employ a kinematic bicycle model \cite{lavalle2006planning} to simulate the dynamics of a four-wheeled vehicle, by positioning the front wheel at the center of the front axle and the rear wheel at the center of the rear axle. The steering angle $\delta$ constrains the vehicle to drive on a circle with a minimum radius $R = L/\tan(\delta)$, where $L$ is the wheelbase. For AVs' path planning problem, we establish a configuration-space (C-space), denoted as $ \chi \subset \mathbb{R}^n $, within which the curvature of the road in the vehicle's C-space is quantified by $\kappa = 1/R$.

\begin{figure}[htbp]
    \centering
    \includegraphics[width=0.4\columnwidth]{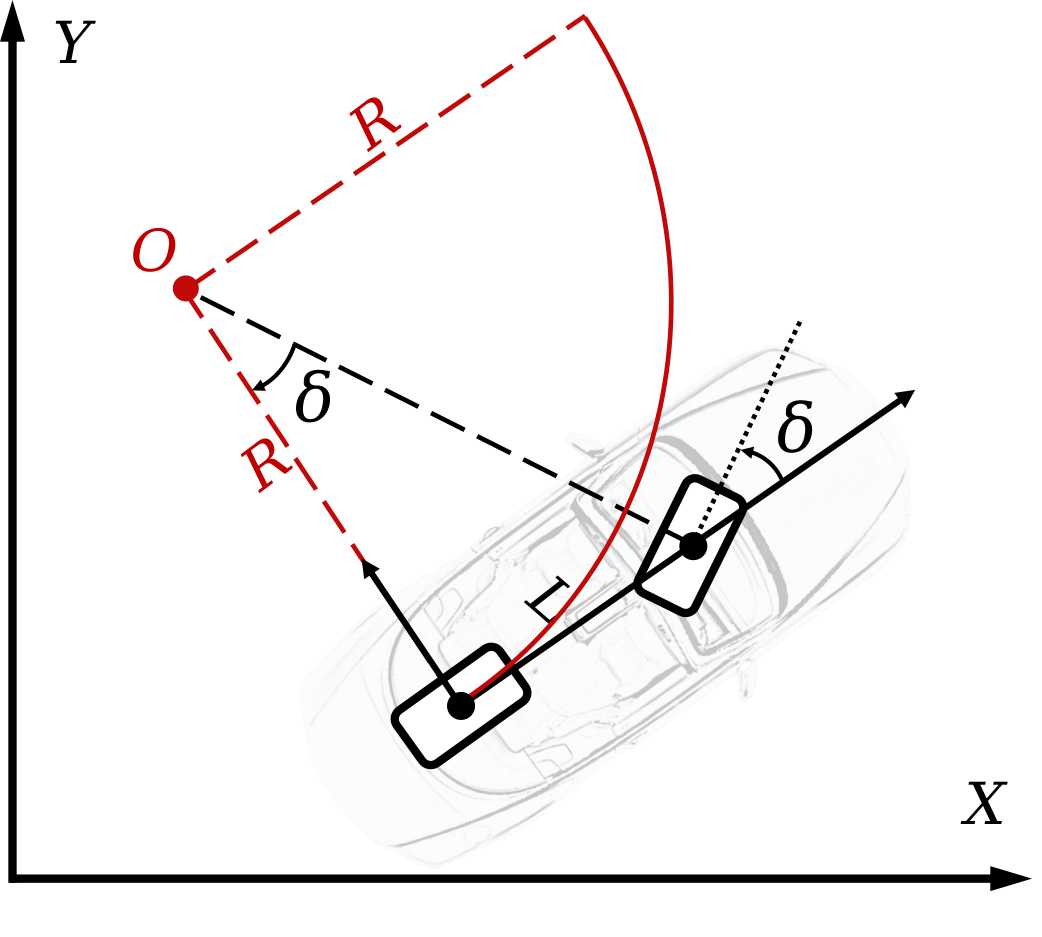}
	\caption{Illustration of the bicycle model.}
	\label{fig:bicycle}
\end{figure}

\begin{figure}[htbp]
    \centering
    \includegraphics[width=0.75\columnwidth]{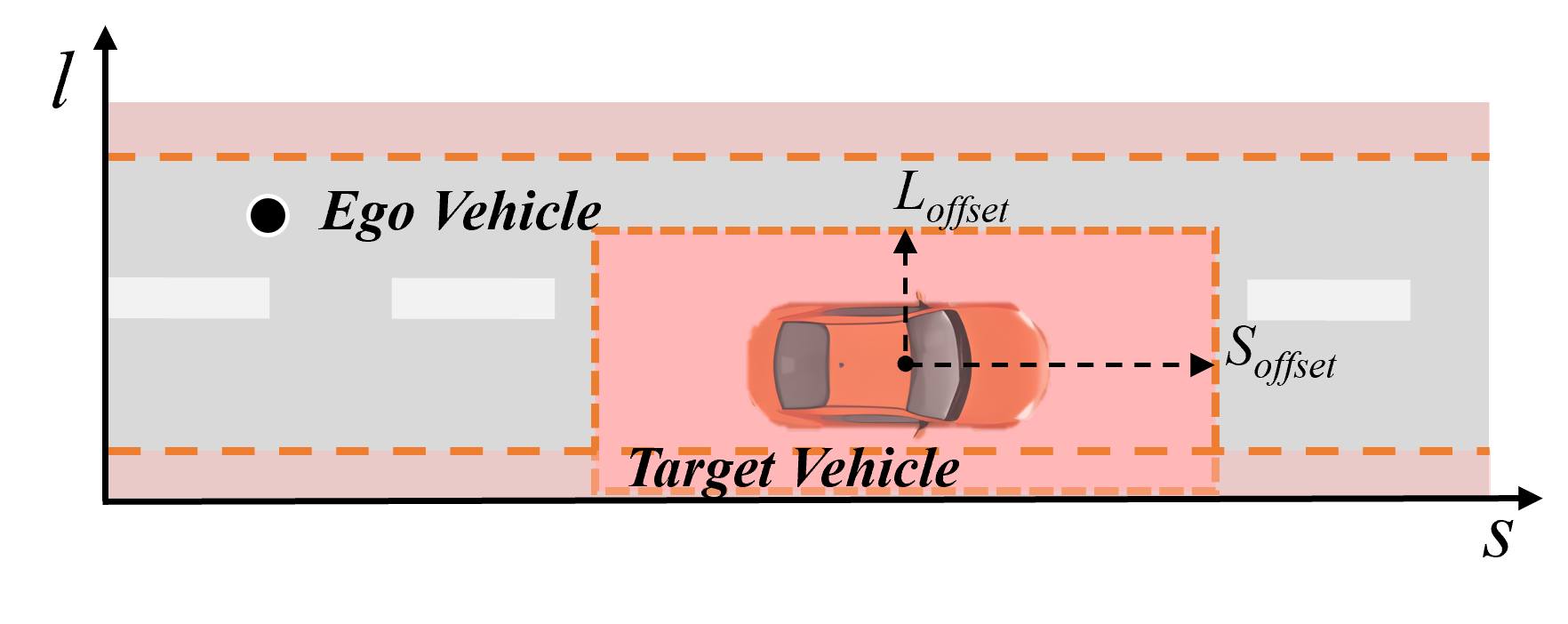}
	\caption{Illustration of configuration space}
	\label{fig:rec}
\end{figure}

The Frenét frame is adopted in trajectory planning because it effectively models structured environments and traffic behaviors \cite{Werling.03.05.201007.05.2010}. Typically, this frame separates spatial dimensions into two orthogonal axes, $s$ and $l$, which allows for decoupled longitudinal and lateral descriptions of the motion of the ego vehicle and other observed traffic entities. In the C-space as depicted in Fig. \ref{fig:rec}, the ego vehicle is modeled as a mass point, with its width establishing permissible boundaries in the Frenét frame. Additionally, the spatial occupation of surrounding vehicles is expanded, with longitudinal dimension $S_\textrm{offset}$ and lateral dimension $L_\textrm{offset}$.

\subsection{B-spline Curve and B-spline Trajectory}\label{sec:B-spline}

In trajectory planning, B-splines are fundamental for crafting smooth and adaptable trajectory for guiding autonomous systems through dynamic environments. Central to the utility of B-splines are several key properties, including their convex hull property, which ensures that the curve lies within the convex hull formed by its control points \cite{piegl2012nurbs}. 

B-splines are defined by a set of control points, denoted as $Q_i$, which represent the waypoints of the trajectory, and a knot vector, denoted as $T = \{t_0, t_1, ... , t_{M-1}\}$, where $M = N_c + p_b + 1$, with $N_c$ being the number of control points and $p_b$ the degree of the B-spline curve. The knot vector determines the parameterization of the curve and influences its shape. The B-spline curve $\Phi (t)$ of degree $p_b$ with $N_c$ control points is expressed as \cite{piegl2012nurbs}:

\begin{equation}
\Phi(t) = \sum_{i=0}^{N_c-1} Q_{i}B_{i,p_b}(t),
\end{equation}
where $B_{i,p_b}(t)$ are the B-spline basis functions. These basis functions are piecewise polynomial functions defined recursively. One common formulation of these basis functions is the Cox-de Boor recursion formula \cite{piegl2012nurbs}:

\begin{equation}
B_{i,0}(t) = \begin{cases} 
1 & \text{if } t_i \leq t < t_{i+1} \\
0 & \text{otherwise}
\end{cases}
\end{equation}

\begin{equation} \label{eq: bspline}
B_{i,p_b}(t) = \frac{t - t_i}{t_{i+p_b} - t_i} B_{i,p_b-1}(t) + \frac{t_{i+p_b+1} - t}{t_{i+p_b+1} - t_{i+1}} B_{i+1,p_b-1}(t),
\end{equation}

for $p_b > 0$. \\

A special type of B-spline is called uniform B-spline. Each knot of a uniform B-spline is separated by the same time interval $\delta t=t_{i+1}-t_i$.
Uniform B-splines offer engineers a straightforward framework for trajectory planning, allowing for precise control over the trajectory while maintaining desirable properties such as continuity, smoothness, and adherence to the convex hull formed by the control points. Furthermore, uniform B-spline curves are continuously differentiable up to $p_b - 1$ times, ensuring smooth motion profiles suitable for various robotic applications \cite{EGO2021Zhou}.
The control points of the velocity ${V}_i$, acceleration ${A}_i$, and jerk ${J}_i$ curves therefore can be obtained by \cite{EGO2021Zhou}
    \begin{equation}
    \label{equ:v_a_j}
    {V}_i=\frac{{Q}_{i+1}-{Q}_i}{\triangle t}, \ {A}_i=\frac{{V}_{i+1}-{V}_i}{\triangle t}, \ {J}_i=\frac{{A}_{i+1}-{A}_i}{\triangle t}.
    \end{equation}

\section{Safety Metric Aware Trajectory Repairing} \label{sec:Method}
\subsection{Approximating the Feasible Time-to-React}

We begin by providing some essential definitions. 

\textbf{Definition 1 (Time-To-React):} \textit{The Time-To-React (TTR) is the maximum duration the ego vehicle can follow the reference trajectory $u([t_0,t_h])$ without resulting in a collision. The reference trajectory $u([t_0,t_h])$ is generated by a nominal trajectory planner over the time span $[t_0,t_h]$. Here, $t_0$ represents the initial time, and $t_h$ denotes the time horizon for the reference trajectory.}

In previous work \cite{tong2023robust, lin2021sampling}, the TTR was estimated using a fixed evasive trajectory (full deceleration, kick-down, full-steering, etc.); however, this trajectory is not the one that the controller will actually follow. Furthermore, it does not ensure a collision-free interaction with all traffic participants or dynamic feasibility. Therefore, we propose the Feasible Time-to-React (F-TTR), which is calculated based on a feasible trajectory that will be used by trajectory tracking controllers.

\textbf{Definition 2 (Feasible Time-To-React:} \textit{The Feasible Time-To-React (F-TTR) is the maximum amount of time in which the ego vehicle can adhere to the reference trajectory $\Phi_r([t_0,t_h])$ with respect to variable $t$ and execute the appropriate trajectory created by the trajectory planner to avoid a possible collision. The feasible trajectory is free of collisions, dynamically feasible, and will be tracked by a low-level controller.}

To approximate the F-TTR, we propose a binary search algorithm, as detailed in Algorithm \ref{alg:f-ttr}.
Based on the reference trajectory and trajectories of all traffic participants, the algorithm foremost estimates the TTC using detectCollision(-) (line 1) and the feasible repaired trajectory $\Gamma(t)$ is initialized with the reference trajectory $\Phi_r(t)$ (line 2). The F-TTR is set to 0 if a collision is already detected (line 4). If no collision is detected, the F-TTR is considered to be infinite (line 5).
In all other cases, a binary search algorithm determines the F-TTR within the constraints of step difference and the function inTimeLimit() (from line 7 to line 17). Ultimately, the F-TTR and feasible repaired trajectory is returned. 
In line 12, the isFeasible() function utilizes the CommonRoad Drivability Checker \cite{pek2020commonroad} to assess the feasibility of the generated trajectory.

\begin{algorithm}
\caption{Binary Search for F-TTR}\label{alg:f-ttr}
\begin{algorithmic}[1] 
\Require $P_0$: Set of the reference trajectory $\Phi_r(t)$ and predicted trajectories of other vehicles,  $\mathscr{T}(P_0, t_{\text{rep}, c})$: a trajectory planner with respect to $P_0$ , time-to-repair $t_{\text{rep}}$ and constraints $c$, $\Delta T$: time resolution for binary search
\State ${\text{TTC}} \gets \text{detectCollision}(P_0)$
\State $\Gamma(t) \gets \Phi_r(t)$
\If{$\text{TTC} == 0$}
    \State $\text{F-TTR} \gets 0$
\ElsIf{$\text{TTC} == \infty$}
    \State $\text{F-TTR} \gets \infty$
\Else
    \State $t_{\text{rep}} = t_{\text{start}} \gets 0$
    \State $t_{\text{end}} \gets \text{TTC}$
    \While{$| t_{\text{end}} - t_{\text{start}} | > \Delta T$ and $\text{inTimeLimit}()$}
        \State ${\Phi}_{t \geq t_\text{rep}}(t) \gets \mathscr{T}(P_0, t_{\text{rep}})$
        \If{$\text{isFeasible}({\Phi}_{t \geq t_\text{rep}}(t))$}
            \State $t_{\text{start}} \gets t_{\text{rep}}$
            \State $\Gamma(t) \gets {\Phi}_{t \geq t_\text{rep}}(t)$
        \Else
            \State $t_{\text{end}} \gets t_{\text{rep}}$
        \EndIf
         \State $t_{\text{rep}} \gets (t_{\text{start}} + t_{\text{end}}) / 2$ 
    \EndWhile
    \State $\text{F-TTR} \gets t_{\text{rep}}$
\EndIf
\State \Return $\text{F-TTR}$, $\Gamma(t)$
\end{algorithmic}
\end{algorithm}

\begin{figure}[t] 
    \centering
    \includegraphics[width=1.0\columnwidth]{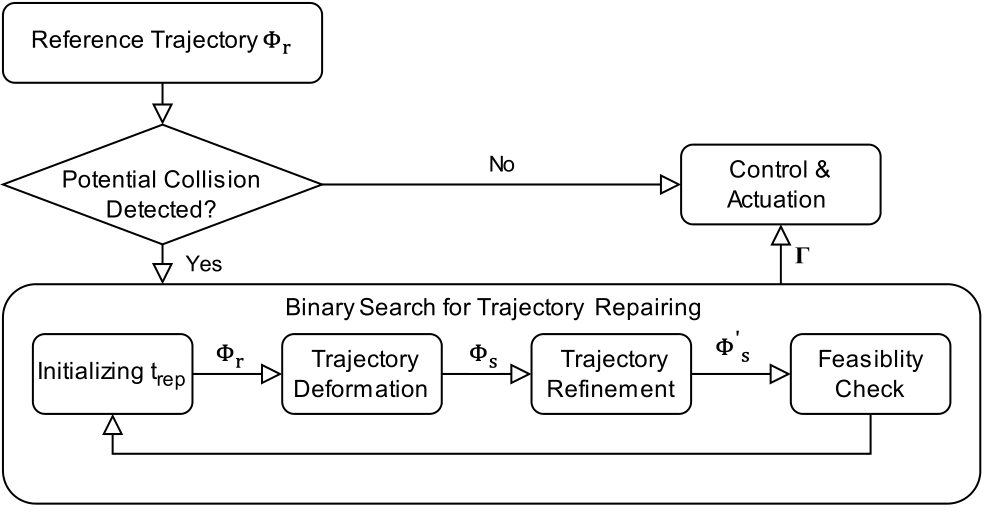}
	\caption{Flowchart of the proposed trajectory repairing framework 
}
	\label{fig:overview}
\end{figure}

The flowchart in Fig. \ref{fig:overview} outlines our proposed process for repairing a reference trajectory, applicable to both AVs and robotics. The process begins with a reference trajectory, denoted as \(\Phi_r(t)\), which is then assessed for potential collisions. If no collisions are detected, the system continues to follow the planned trajectory using control and actuation mechanisms. However, if a potential collision is identified, the system employs a binary search algorithm (Algorithm \ref{alg:f-ttr}) to determine the F-TTR and identify a feasible repaired trajectory. 

This algorithm begins by setting a time variable \( t_{\text{rep}} \) and then deforms the reference trajectory \( \Phi_r(t) \) into \( \Phi_s(t) \) to avoid collisions. The deformed trajectory \( \Phi_s(t) \) is then further refined to \( \Phi'_s(t) \).
A feasibility check on this refined trajectory confirms its feasibility, incorporating safety considerations and physical limits. So that the final repaired trajectory $\Gamma(t)$ satisfies the vehicle dynamics and safety constraints.  Based on the results of the feasibility check, \(t_\text{rep}\) is reinitialized as depicted in Algorithm \ref{alg:f-ttr}. Ultimately, the driving system implements the repaired trajectory with a safety metric (F-TTR), which enables the vehicle to avoid potential collisions and allows for non-immediate responses.
 
\subsection{Trajectory Repairing Using B-spline Curve Optimization}
The problem formulation in this research builds on the sophisticated Ego-Planner framework for quadrotor local planning \cite{EGO2021Zhou}.The reference trajectory is defined by a uniform B-spline curve $\Phi_{r}(t)$, the final repaired feasible trajectory is also an uniform B-spline curve denoted as  $\Gamma(t)$.   This curve is defined by its degree \(p_b\), a sequence of control points \(\{Q_0, Q_1, Q_2, \ldots, Q_{N_c-1}\}\), and a corresponding knot vector \(\{t_0, t_1, t_2, \ldots, t_{M-1}\}\). Each control point \(Q_i \in \mathbb{R}^2 \), which is the two-dimensional Frenét frame, 
and each knot \(t_m \in \mathbb{R} \).

\begin{figure}[bh] 
    \centering
    \includegraphics[width=0.8\columnwidth]{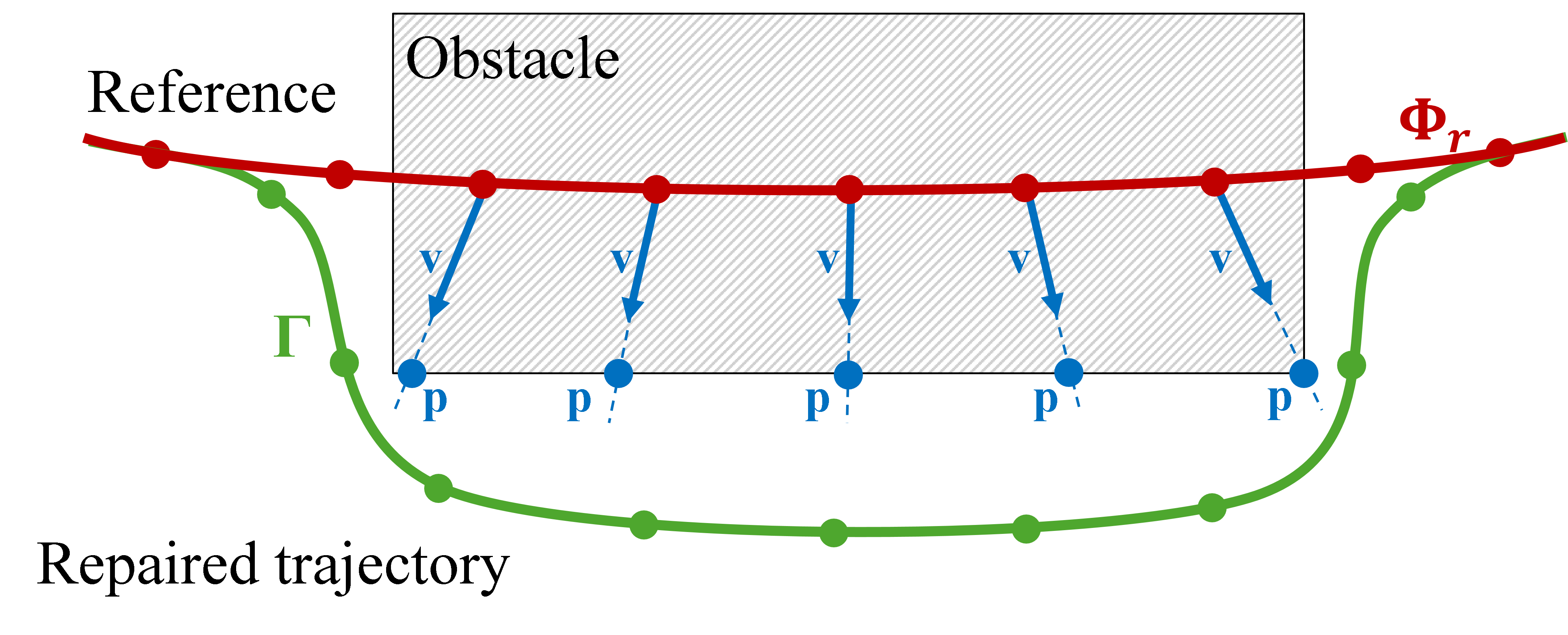}
	\caption{Illustration of the trajectory deformation and refinement using B-spline }
	\label{fig:repair}
\end{figure}

Our approach to trajectory repairing is structured around the optimization of B-spline curves, and it is divided into two primary phases: trajectory deformation and trajectory refinement.  To facilitate a clear understanding of our methodology, we introduce the related definitions in the following part.

\textbf{Definition 3 (Obstacle Distance \cite{EGO2021Zhou}):} \textit{If a potential collision is detected along the reference trajectory, consider a control point $Q_i$ that forms a trajectory colliding with the $j^\text{th}$ obstacle. The distance from $Q_i$ to the $j^\text{th}$ obstacle, denoted as $d_{ij}$, is defined by the equation}
\begin{equation}
    d_{ij} = (Q_i - p_{ij}) \cdot v_{ij}
\end{equation}
\textit{where $p_{ij} \in \mathbb{R}^2$ represents an anchor point on the surface of the obstacle (including a safety margin) and $v_{ij} \in \mathbb{R}^2$ is the unit vector in the direction from $Q_i$ to $p_{ij}$. This configuration is illustrated in Figure \ref{fig:repair}.}

The obstacle distance will be further used in trajectory deformation.
\subsection{Trajectory Deformation}
The optimization problem for trajectory deformation is formulated as follows \cite{EGO2021Zhou}:
\begin{equation}
    \min_{Q} J = \lambda_s J_s + \lambda_c J_c + \lambda_d J_d
\end{equation}
where the penalty function $J_s$ represents the penalty for \textbf{smoothness}, $J_c$ represents the penalty for \textbf{collision}, and $J_d$ shows the penalty for \textbf{feasibility}.

\textbf{Smoothness penalty function:} The smoothness penalty function reduces the magnitude of higher-order derivatives, resulting in a smoother trajectory, which is formulated as follows:
\begin{equation}
    J_s = \sum_{i = 0}^{N_c - 3} \|A_i\|_2^2 + \sum_{i = 0}^{N_c - 4} \|J_i\|_2^2
\end{equation}

\textbf{Collision penalty function:}   The collision penalty forces the control points to move away from collisions.
    This is accomplished by implementing a safety threshold, denoted as $s_f$, and penalizing control points that have a distance $d_{ij}$ less than the safety threshold.
    We utilize a penalty function $j_c$ that is twice continuously differentiable, which is identical to the one used in \cite{EGO2021Zhou}. The function is defined as: 

\begin{equation}
    \begin{aligned}
        j_c(i, j) & =\left\{\begin{array}{lr}
        0 & \left(c_{i j} \leq 0\right) \\
        c_{i j}^3 & \left(0<c_{i j} \leq s_f\right) \\
        3 s_f c_{i j}^2-3 s_f^2 c_{i j}+s_f^3 & \left(c_{i j}>s_f\right)
        \end{array}\right. \\
        c_{i j} & =s_f-d_{i j}
    \end{aligned}
\end{equation}
where the cost value created by the pairs $\{{p},{v}\}_j$ on ${Q}_i$ is denoted as $j_c(i,j)$.
The summation of costs on all ${Q}_i$ results in the overall cost $J_c$, which gets expressed as:
 \begin{equation}
    J_c = \sum_{i = 0}^{N_c-1} j_c (Q_i)
\end{equation}

\textbf{Feasibility penalty function:}
Feasibility is assured by constraining the higher order derivatives of the trajectory on every single dimension, by applying $|{\Phi}^{(k)}_{s, \gamma}(t)| < {\Phi}^{(k)}_{\gamma,max}$ for every $t$, where $\gamma \in \{s, l\}$ represents each dimension in the Frenét frame and $ {\Phi}^{(k)}_{\gamma,max}$ is the physical limit. 
Therefore, by utilizing the convex hull feature, the penalty function is defined as
\begin{equation}
    J_d=\sum_{i=0}^{N_c-2} w_v F\left({V}_i\right)+\sum_{i=0}^{N_c-3} w_a F\left({A}_i\right)+\sum_{i=0}^{N_c-4} w_j F\left({J}_i\right)
\end{equation}
Here, $w_v$, $w_a$, and $w_j$ represent the weights assigned to each term, and $F(\cdot)$ is a metric function that is two times continuously differentiable and depends on higher order derivatives of control points. The function $F(\cdot)$ is defined in accordance with the definition provided in the reference \cite{EGO2021Zhou} as follows:

\begin{equation}
    F({C})=\sum_{r=s, l} f\left(c_r\right)
\end{equation}
with 
\begin{equation}
f\left(c_r\right)=\left\{\begin{array}{lr}
a_1 c_r^2+b_1 c_r+c_1 & \left(c_r \leq-c_j\right) \\
\left(-\lambda c_m-c_r\right)^3 & \left(-c_j<c_r<-\lambda c_m\right) \\
0 & \left(-\lambda c_m \leq c_r \leq \lambda c_m\right) \\
\left(c_r-\lambda c_m\right)^3 & \left(\lambda c_m<c_r<c_j\right) \\
a_2 c_r^2+b_2 c_r+c_2 & \left(c_r \geq c_j\right)
\end{array}\right.
\end{equation}
where the variable $c_r \in {C} = \{{V}_{i}, {A}_{i}, {J}_{i}\}$. The coefficients \(a_1\), \(b_1\), \(c_1\), \(a_2\), \(b_2\), and \(c_2\) are specifically chosen to ensure second-order continuity of the B-spline curve, adhering to the conditions stated in \cite{EGO2021Zhou}.
The parameter \(c_m\) denotes the upper limit of the derivative, critical for maintaining the dynamic feasibility of the trajectory, while \(c_j\) indicates the points of transition between quadratic and cubic segments of the trajectory curve. 
Additionally, the coefficient \(\lambda\), constrained by \(\lambda < 1 - \epsilon\) (where \(\epsilon \ll 1\)), acts as an elastic coefficient. The role of \(\epsilon\) is to fine-tune the balance within the cost function, facilitating an optimal trade-off among several weighted terms involved in the optimization process \cite{EGO2021Zhou}.

\subsection{Trajectory Refinement}
The trajectory deformation process accounts for feasibility, however, the optimization operates under soft constraints, potentially exceeding physical limits. Also, the primary objective during this phase is to avoid obstacles, which may result in a trajectory that is not smooth.

Consequently, it becomes essential to re-optimize the safe trajectory, \( \Phi_s \), from the previous step. This re-optimization process regenerates \( \Phi_s \) into a refined trajectory \( \Phi'_s \). The optimization is guided by a penalty function \( J' \), which is a linear combination of the \textbf{smoothness} penalty function \( J_s \), the \textbf{feasibility} penalty function \( J_d \), and the \textbf{curve fitting} penalty function \( J_f \). This is formulated as follows:

\begin{equation}
    \min_{Q} J' = \lambda_s J_s + \lambda_d J_d + \lambda_f J_f
\end{equation}
where \(\lambda_f\) is the weighting factor for the curve fitting penalty function \( J_f \).

The penalty function \( J_f \) is calculated by integrating the anisotropic displacements between points on  \(\Phi_s(\alpha T)\) and the corresponding points on \(\Phi'_s(\alpha T')\)  . Here, \( T \) and \( T' \) represent the durations of the trajectories \(\Phi_s\) and \(\Phi'_s\), respectively, and \(\alpha\) is a parameter that ranges from 0 to 1. For a comprehensive explanation of the fitting penalty function, please see the reference \cite{EGO2021Zhou}.

\section{Evaluation}\label{sec:eval}
We evaluate our approach by using traffic scenarios provided from the open-source CommonRoad platform \cite{althoff2017commonroad}. The computational framework is coded in Python and runs on a PC equipped with an Intel Core i7-12700H processor. The reference trajectory is created using a search-based planner, as described in \cite{tong}. The vehicle characteristics for the ego vehicle are configured to align with the specifications of a Ford Escort, as described in further detail in the reference \cite{althoff2017commonroad}.

\textbf{Numerical optimization:} we leverage the SciPy library and especially employ the L-BFGS solver. This solver is a quasi-Newton approach noted for its effectiveness in solving large-scale optimization problems. The L-BFGS algorithm offers many advantages by using second-order Taylor expansions to estimate the curvature of the objective function. This approach significantly improves both the accuracy and speed of convergence \cite{EGO2021Zhou}. The maximum number of iterations for the L-BFGS algorithm is set to 100, and the tolerance is set to 0.01.

\subsection{Baseline: Quadratic Programming Trajectory Repairing}
In our previous research \cite{tong2023robust}, we used Bernstein basis polynomials and path-speed decoupling in the Quadratic Programming Trajectory Repairing (QPTR) approach to effectively prevent collisions with high computation efficiency. Nevertheless, the issue lies in the fact that the calculation of safety metrics and the construction of an evasive trajectory are not integrated, resulting in inefficiency and the possibility of an unachievable safety measure. We developed our baseline approach following \cite{tong2023robust} and further enhanced it based on the insights from \cite{Deolasee.2023}. In the current implementation of QPTR, both path and speed are integrated into a spatio-temporal framework.

\begin{table}[htbp]
    \centering
    \caption{Comparison of approximated TTR and trajectory repairing results in Scenario 1 and 2. Max lon. dec. represents maximal longitudinal deceleration, while max lat. acc. denotes maximal lateral acceleration.}
    \label{tab:ttr} 
    \begin{tabular}{@{}lcccc@{}} 
        \toprule
        & \multicolumn{2}{c}{Scenario 1 (TTC = 2.9 s)} & \multicolumn{2}{c}{Scenario 2 (TTC = 1.6 s)} \\
        \cmidrule(lr){2-3} \cmidrule(lr){4-5} 
        & (F-)TTR & Max lon. dec. & (F-)TTR & Max lat. acc. \\
        \midrule
        QPTR \cite{tong2023robust} & 2.6 s & not solvable & 1.3 s & infeasible \\
        Our & 1.6 s & -1.86 $m/s^2$ & 0.8 s & 4.82 $m/s^2$ \\
        \bottomrule
    \end{tabular}
\end{table}

\subsection{Scenario 1: Urban T-Intersection}
\begin{figure*}[htbp]
    \centering
    \includegraphics[width=1.0\textwidth]{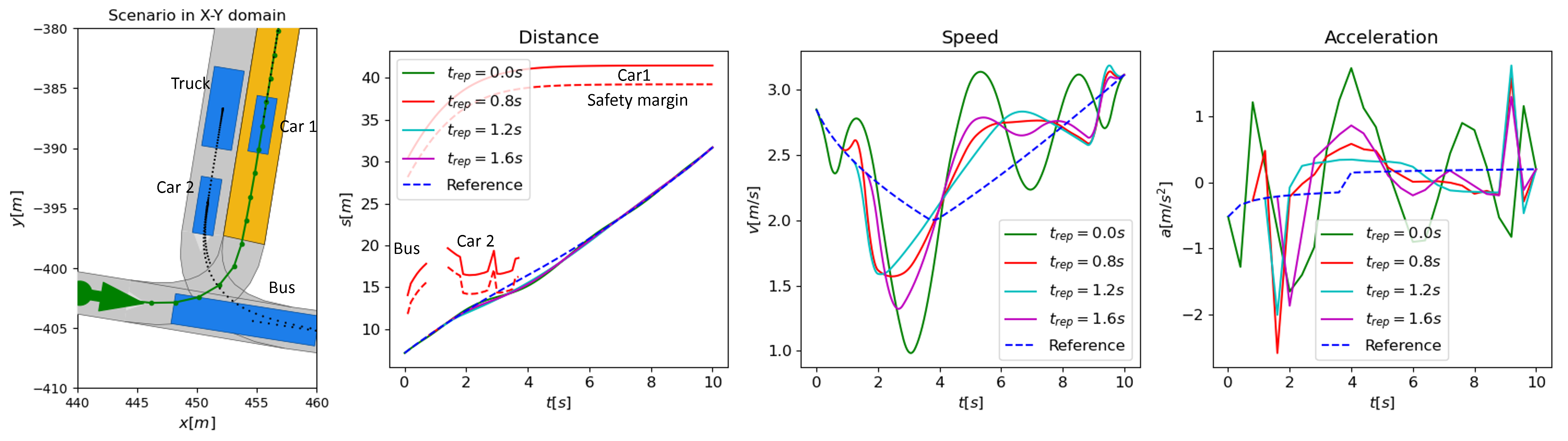}
	\caption{The first diagram illustrates an urban T-intersection scenario, where the planning process starts with the arrow tracing a green line and is dedicated to reach the yellow target zone. The second image on the left shows the extreme points of the other vehicles projected onto the S-T domain (in the ego lane), along with the results of the trajectory repairing. The third and fourth figures illustrate the outcomes of the repairing for speed and acceleration by using different values of $t_\text{rep}$ in the binary search.}
	\label{fig:scenario_1_repairing}
\end{figure*}

We use a complex urban T-intersection scenario, identified as CommonRoad ID: \texttt{DEU\_Flensburg\_6\_1\_T-1}, to verify the effectiveness of our method. The scenario animation is included inside the Scenario Selection Tool of CommonRoad. Based on the flowchart shown in Figure \ref{fig:overview}, our first step involves calculating the TTC, which amounts to 2.9 seconds. Car 2's safety margin is violated by the reference trajectory, see Fig. \ref{fig:scenario_1_repairing}. It is believed that each automobile in the scenario has a rectangle form.  We calculate the predicted trajectories of the outermost points of obstacles, taking into account the safety margin, inside the space-time domain of the ego-driving lane. Two automobiles are seen crossing the driving lane in the center of Figure \ref{fig:scenario_1_repairing}.

In the simulation, the longitudinal safety margin \(S_{\text{offset}} = 2.25  \, \text{m}\) accounts for the length of the ego vehicle. The lateral offset is \(L_{\text{offset}} = 2.0  \, \text{m}\) and the clearance \(s_f = 1.0  \, \text{m}\). The optimization weights are set as follows: \(\lambda_s = 1.0\), \(\lambda_c = 15.0\), \(\lambda_d = 1.0\) for the trajectory deformation, and \(\lambda_s = 1.0\), \(\lambda_d = 1.0\), \(\lambda_f = 0.01\) for the trajectory refinement. The search time resolution \( \Delta T \) is \( 0.4  \, \text{s} \).

The binary search consists of four iterations, occurring in the following sequence: \(t_{\text{rep}} = 0.0 \, \text{s}, 0.8 \, \text{s}, 1.2 \, \text{s}, 1.6 \, \text{s}\).The search stops when the search step difference is smaller than a resolution of 0.4s. In this case, the F-TTR is 1.6 seconds, as seen in Table \ref{tab:ttr}. As the value of $t_\text{rep}$ increases, the minimum velocity decreases, requiring the cars to decelerate more. However, this also allows the ego vehicle to have more reaction time to verify the behavior of other vehicles.
Fig. \ref{fig:scenario_1_repairing} illustrates the trajectory refinement results compared to the trajectory deformation results for \(t_\text{rep}=1.6 \, s\). It is evident that the trajectory refinement process effectively minimizes and smooths the acceleration and speed profiles from the trajectory deformation stage, thereby enhancing the overall outcome.

In contrast, the QPTR approach employs evasive maneuvers to determine the TTR, which is calculated as 2.6 seconds, as shown in Table \ref{tab:ttr}. However, the formulated optimization problem for QPTR with the TTR is not solvable. This issue arises because QPTR must first generate a convex driving corridor by approximating the collision-free space using piecewise linear functions. Unfortunately, this linear approximation is more conservative compared to the actual collision-free space. In a very narrow solution space, such as when repairing critically begins at the TTR in scenario 1, a feasible convex driving corridor does not exist. 
This also indicates that decoupling the search for the TTR and the trajectory repairing process might lead to an unsolvable trajectory repairing problem.

\begin{figure*}[htp]
    \centering
    \includegraphics[width=0.8\textwidth]{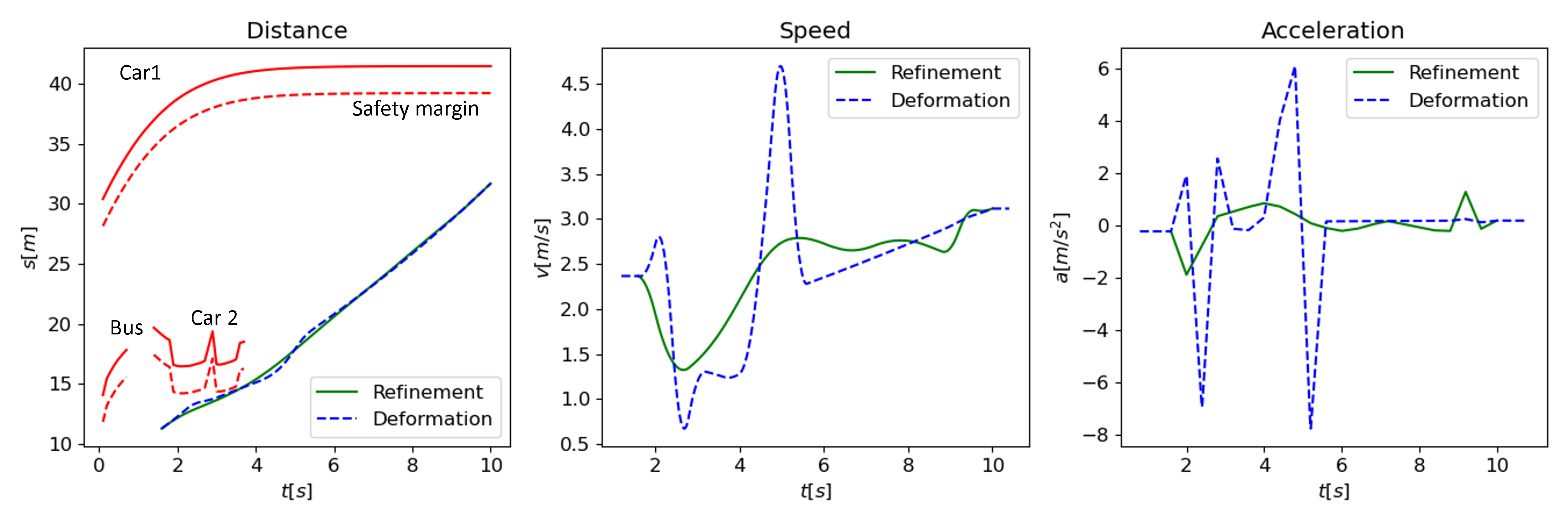}
	\caption{Refined distance, speed, and acceleration at $t_\text{rep} = 1.2 s$. Blue dashed lines indicate the trajectory deformation outcomes, while green lines indicate the trajectory refinement outcomes.}
	\label{fig:deformation_refinement}
\end{figure*}

\begin{figure*}[htbp]
    \centering
    \includegraphics[width=0.8\textwidth]{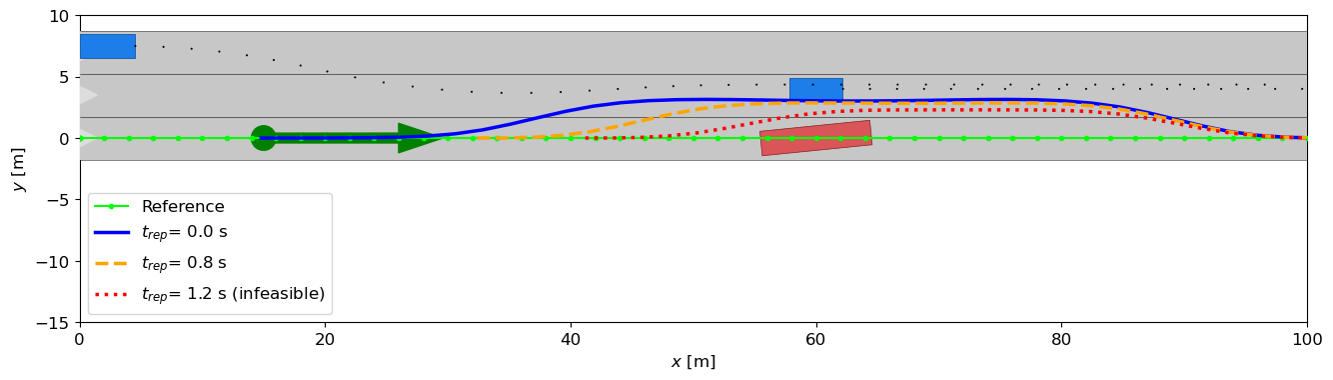}
	\caption{Scenario 2 trajectory repairing results. The planning begins with the arrow following a green reference path. The dynamic obstacles are in blue, while the road damage (static obstacle) is in red. The dotted black lines represent the future motion of other vehicles.  }
	\label{fig:scenario_2_repairing}
\end{figure*}

\subsection{Scenario 2: Road Damage Avoidance in Dynamic Traffic}
The second scenario was inspired by the EU-H2020-funded project ESRIUM \cite{Rudigier2022}, where a digital map for road deterioration is created.
We have developed a duplicate of the scenario.
As seen in Fig. \ref{fig:scenario_2_repairing}, the ego vehicle must change lanes to evade road damage. Nevertheless, there is a car in motion in the next lane, and another car in the third lane about to change lanes into the second lane. This poses a hurdle for the trajectory planner.

Fig.~\ref{fig:scenario_2_repairing} presents a visual comparison of multiple results of repaired trajectories, each with different values of \( t_{\text{rep}} \), in the presence of dynamic situations. The parameters used for the numerical experiments are as follows: \( S_{\text{offset}} = 8.0 \),  \( L_{\text{offset}} = 2.0  \, \text{m} \) and \( s_f = 1.3  \, \text{m} \). For the trajectory deformation, the weights are: \( \lambda_s = 1.0 \), \( \lambda_c = 12.0 \), \( \lambda_d = 0.5 \). For trajectory refinement, the weights are \( \lambda_s = 1.0 \), \( \lambda_d = 10.0 \), \( \lambda_f = 0.01 \). The search time resolution \( \Delta T \) is \( 0.4  \, \text{s} \). After the proposed binary search, the F-TTR is 0.8 seconds with a maximal lateral acceleration of $4.82 m/s^2$ (Table \ref{tab:ttr}). The iteration of repairing terminates at 1.2 seconds, as the generated trajectory is infeasible.

We evaluated the results of QPTR in Scenario 2. It produces a larger TTR (1.3 s), as shown in Table \ref{tab:ttr}. Similar to the issue in Scenario 1, because the repair begins too close to the obstacle, QPTR cannot find a solution that does not exceed the lateral acceleration limit.In contrast, the repaired trajectory from our approach is both dynamically feasible and collision-free with other traffic participants.

Table~\ref{tab:computation_time} presents the computation time of our approach for Scenario 1 and Scenario 2. The longer computation times are attributed to the limitations of the Python implementation and the inherently greater complexity of solving non-linear programming problems compared to quadratic programming. The previous work, which employs quadratic programming and a simplistic TTR search algorithm, consequently has shorter computation time. However, the real-time capability of our approach could be significantly enhanced by utilizing C++ and multi-threading for the optimization, as demonstrated in \cite{zheng2023real}.

\begin{table}[htbp]
\centering 
\caption{Comparison of computation time. We run 100 iterations for each algorithm. The computation time includes the time solving the two-stage optimization problem and feasibility check. The number before and after $\pm$ represent the mean and standard deviation, correspondingly.}\label{tab:computation_time}

\begin{tabular}{ccc}
\hline
Scenario  & Total search time (s) & Search time per iteration (s) \\ \hline
   (1)        & 1.714$\pm$0.004         & 0.426$\pm$0.001                 \\
   (2)       & 0.506$\pm$0.003         & 0.169$\pm$0.001                 \\ \hline
\end{tabular}
\end{table}

\section{Conclusion and Outlook}\label{sec:conclusions}

This work presents an innovative method for repairing trajectories for AVs by directly incorporating safety metrics into the repair process. Our approach aims to preserve the original trajectory as much as possible, maintaining valid segments while selectively repairing invalid sections to safeguard against environmental disturbances. The proposed F-TTR metric offers a more practical safety measure compared to the traditional TTR metric, providing a nuanced understanding of repair needs. This is demonstrated in Table \ref{tab:ttr}, where trajectory repair based on the TTR metric was unsuccessful for QPTR in the presented scenarios, highlighting the limitations of solely relying on traditional metrics. Additionally, we introduce a binary search method for iterative trajectory repair, integrating safety metrics with executable trajectories to ensure the feasibility of the repaired trajectory. By leveraging this approach, we achieve a more resilient trajectory repair mechanism, capable of addressing diverse and dynamic challenges encountered by AVs.

Future enhancements will focus on improving computational performance through the implementation of C++ and multi-threading. Additionally, we plan to extend this framework to robotics applications, specifically for planning the foot trace of legged robots in challenging environments. This expansion holds significant relevance, especially in rough terrain such as in intense Search and Rescue missions.

\addtolength{\textheight}{0cm}   


\bibliographystyle{IEEEtran}
\bibliography{ref.bib}

\end{document}